\documentclass[letterpaper, 10 pt, conference]{ieeeconf}
\IEEEoverridecommandlockouts
\usepackage{amsmath,amssymb,amsfonts}
\usepackage{graphicx}
\usepackage{algorithmic}
\usepackage{acronym}
\usepackage{subcaption}

\makeatletter
\let\NAT@parse\undefined
\makeatother
\usepackage{hyperref}
\usepackage[capitalise]{cleveref}

\title{\LARGE \bf
    Underactuated Waypoint Trajectory Optimization\\for Light Painting Photography
}

\author{Christian Eilers*, Jonas Eschmann*, Robin Menzenbach*,\\
        Boris Belousov, Fabio Muratore, and Jan Peters%
    \thanks{*Authors contributed equally {\tt\scriptsize name.surname@stud.tu-darmstadt.de}}%
    \thanks{All authors are with the Computer Science Department,
        Technische Universit\"{a}t Darmstadt, Germany
        {\tt\scriptsize name.surname@tu-darmstadt.de}}%
    \thanks{This project has received funding from the European Union's Horizon 2020
        research and innovation programme under grant agreement No 640554.}%
}

\begin{document}

\acrodef{rbf}[RBF]{Radial Basis Function}
\acrodef{lqr}[LQR]{Linear Quadratic Regulator}
\acrodef{nlp}[NLP]{Nonlinear Program}
\acrodef{dslr}[DSLR]{Digital Single-Lens Reflex}

\maketitle
\thispagestyle{empty}
\pagestyle{empty}

\begin{abstract}
    Despite their abundance in robotics and nature,
    underactuated systems remain a challenge for control engineering.
	Trajectory optimization provides a generally applicable solution,
	however its efficiency strongly depends on the skill of the engineer
	to frame the problem in an optimizer-friendly way.
	This paper proposes a procedure that automates such problem reformulation for a class of tasks in which the desired trajectory is specified by a sequence of waypoints.
	The approach is based on introducing auxiliary optimization variables that represent waypoint activations.
	To validate the proposed method, a letter drawing task is set up where shapes traced by the tip of a rotary inverted pendulum are visualized using long exposure photography.
\end{abstract}

\section{Introduction}
\label{sec:introduction}
Controlling underactuated systems is of special interest in robotics and engineering
because many common systems such as automobiles, hovercrafts, aircrafts, ships,
legged and wheeled robots, as well as underwater vehicles are underactuated~\cite{spong1998underactuated}.
Nevertheless, designing efficient controllers for such systems requires significantly more effort
than for fully actuated ones~\cite{choset2005principles}.
In particular, even if a feasible trajectory is obtained in simulation,
trajectory tracking on a real system is non-trivial
because not all deviations from the desired trajectory
can be compensated due to the underactuation~\cite{tedrake2009underactuated}.

Although techniques such as partial feedback linearization~\cite{spong1994partial},
which aim to cancel the system dynamics, can be effective at reducing the plant to a partially linear form,
they do not exploit the passive system dynamics~\cite{tedrake2009underactuated}.
For classical control systems, such as the cart-pole, convey-crane, pendubot, etc., a number of controllers have been hand-designed~\cite{fantoni2001non} that do exploit the system dynamics.
The task in those examples is typically to drive the system to an equilibrium state.
In this paper, on the other hand, we are interested in generating and tracking a dynamic trajectory
rather than reaching a static target state.

Trajectory generation for both actuated and underactuated systems is commonly performed using numerical optimization~\cite{tedrake2009underactuated}.
For fully actuated systems, waypoints are relatively straightforward to incorporate into the trajectory generation process because kinematic path planners can be used~\cite{lin2018efficient}.
For underactuated systems, however, a kinematic plan may be dynamically infeasible~\cite{schultz2012trajectory}.
Therefore, a dynamics-based trajectory optimization method is needed that can handle trajectory specification in the form of a sequence of waypoints.

The main contribution of this paper is the design of the objective function for trajectory optimization described in~\cref{sec:planning}.
This objective function explicitly takes into account the waypoints and thus enables the generation of recognizable letter contours in long exposure photography.
However, open-loop execution of the optimal trajectory is not sufficient on its own because even small deviations from the planned trajectory yield unrecognizable letters.
\cref{sec:tracking} details the implementation of the stabilizing feedback controller that enables efficient trajectory tracking.
Finally, the resulting trajectories and letters are presented in~\cref{sec:results}.

\section{Light Painting Setup}
\label{sec:setup}
\begin{figure}[t]
	\centering
	\begin{subfigure}{0.335\columnwidth}
		\centering
		\includegraphics[width=\linewidth]{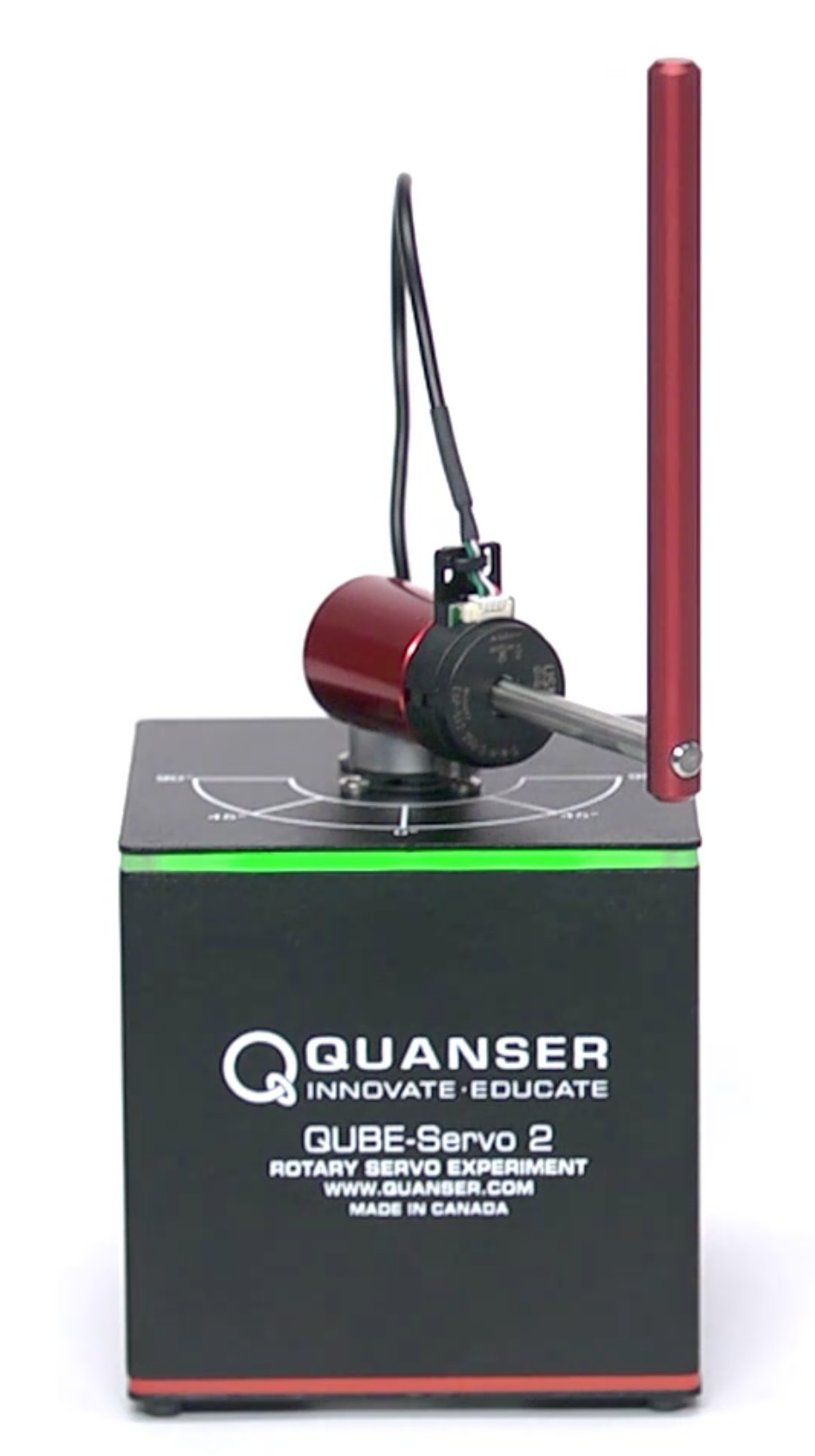}
		\caption{Quanser Qube.}
		\label{fig:qube}
	\end{subfigure}\quad
	\begin{subfigure}{0.6\columnwidth}
		\centering
		\includegraphics[width=\linewidth]{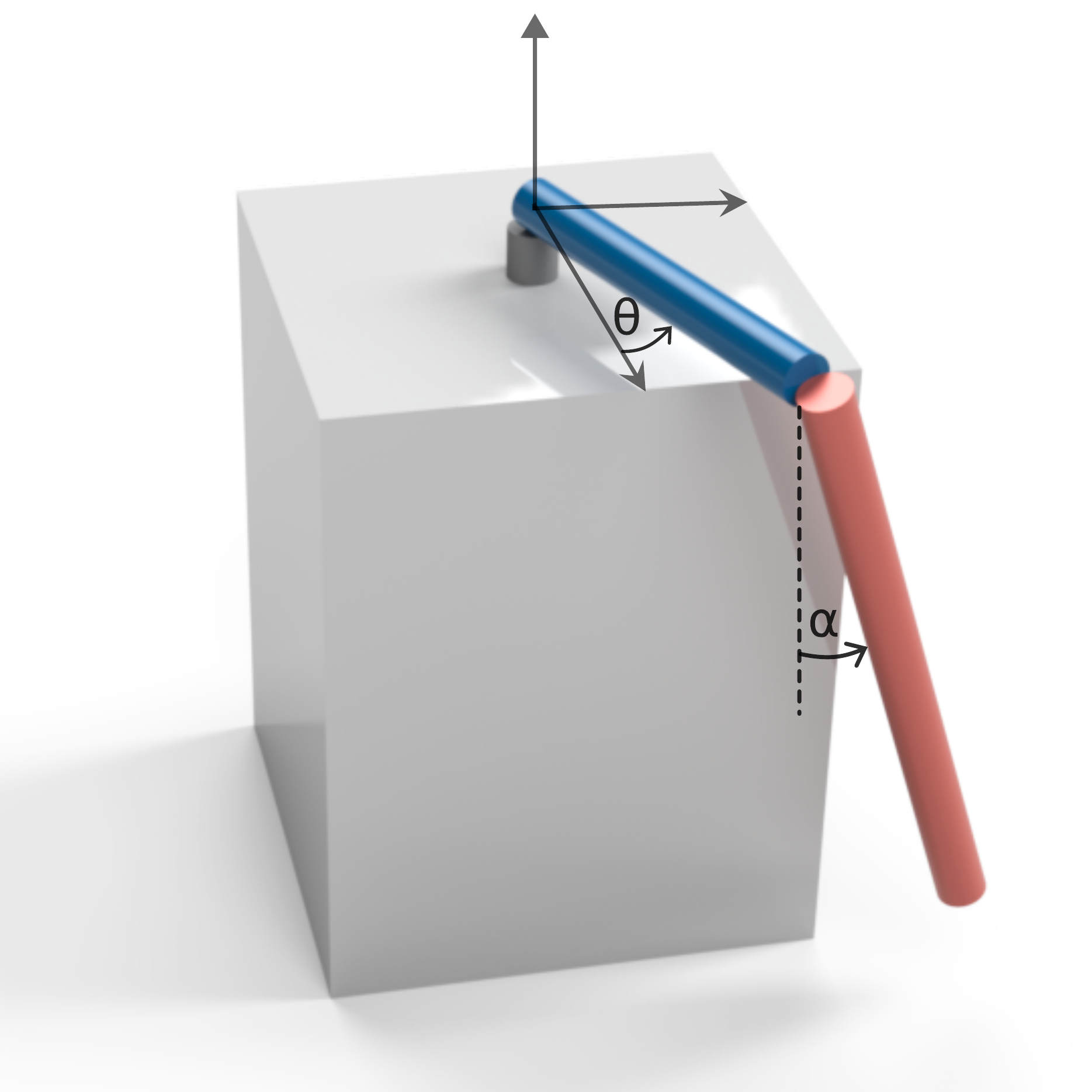}
		\caption{Coordinate systems.}
		\label{fig:furuta}
	\end{subfigure}
	\caption{Evaluation platform: Furuta pendulum.}
	\vspace{-1.0em}
\end{figure}
The hardware platform used for experiments is the Quanser Qube shown in~\cref{fig:qube}.
It implements the rotary inverted pendulum system introduced by Furuta et al.~\cite{furuta1992swing},
which consists of a freely rotating pendulum attached to a motor-driven arm.
A schematic is shown in~\cref{fig:furuta}.
While the arm can be rotated in the horizontal plane,
the pendulum swings in the vertical plane orthogonal to the arm.
The state of the nonlinear system is described by the two angles and the corresponding angular velocities
\begin{equation*}
	\mathbf{x} = \begin{bmatrix}
		\theta & \alpha & \dot{\theta} & \dot{\alpha}
	\end{bmatrix}^T.
\end{equation*}
The Furuta pendulum is a classical platform for evaluating control algorithms,
appreciated for its rich passive dynamics and underactuation.
Its equations of motion are provided in the Appendix,
with derivations starting from the Euler-Lagrange equations
available in~\cite{furuta1992swing}~and~\cite{cazzolato2011dynamics}.

\begin{figure}[t]
	\begin{minipage}{.24\textwidth}
		\centering
		\includegraphics[height=.14\textheight]{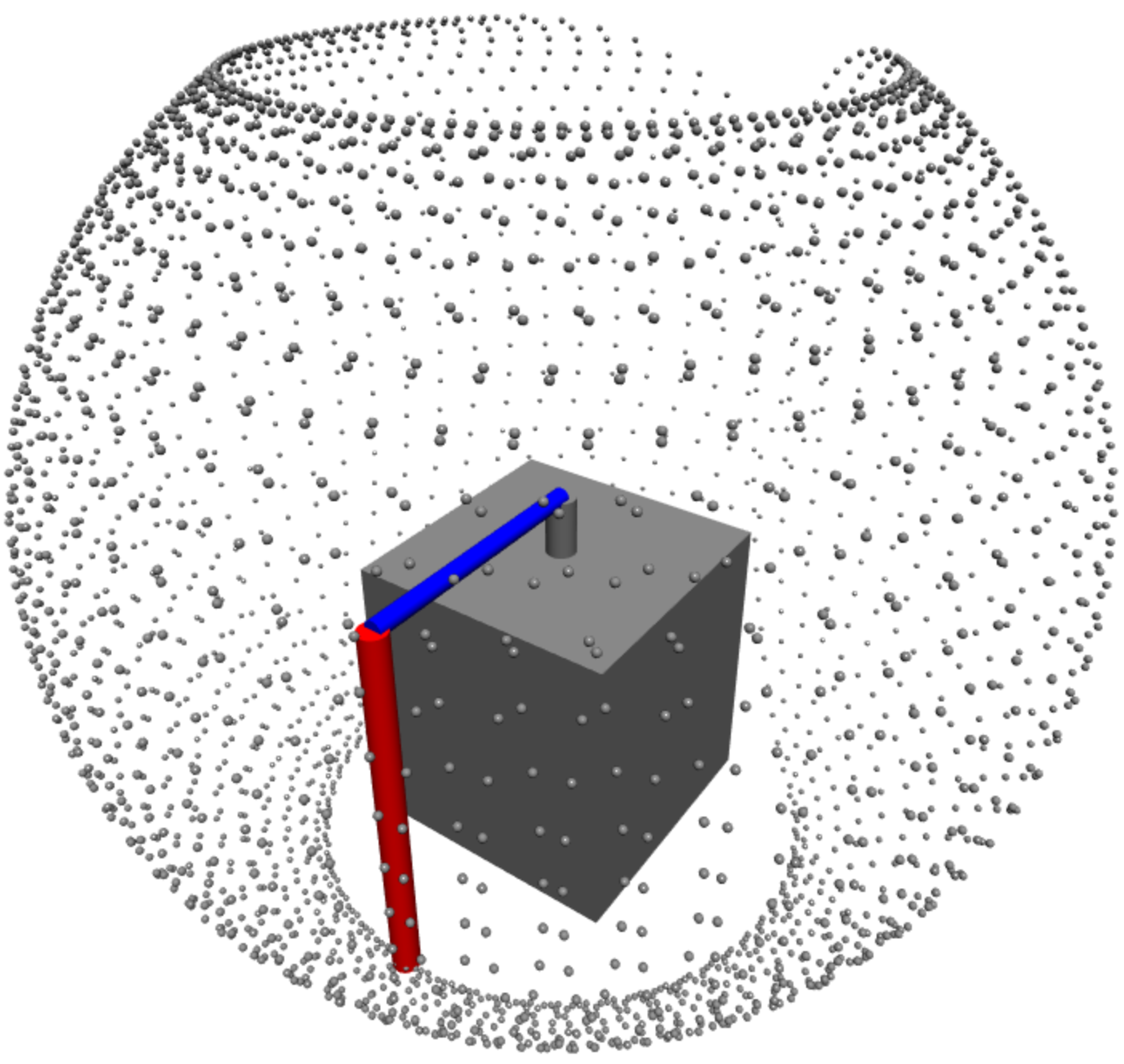}
		\caption{Reachable space\\visualized as a point cloud.}
		\label{fig:reachable}
	\end{minipage}
	\begin{minipage}{.24\textwidth}
		\centering
		\includegraphics[height=.14\textheight]{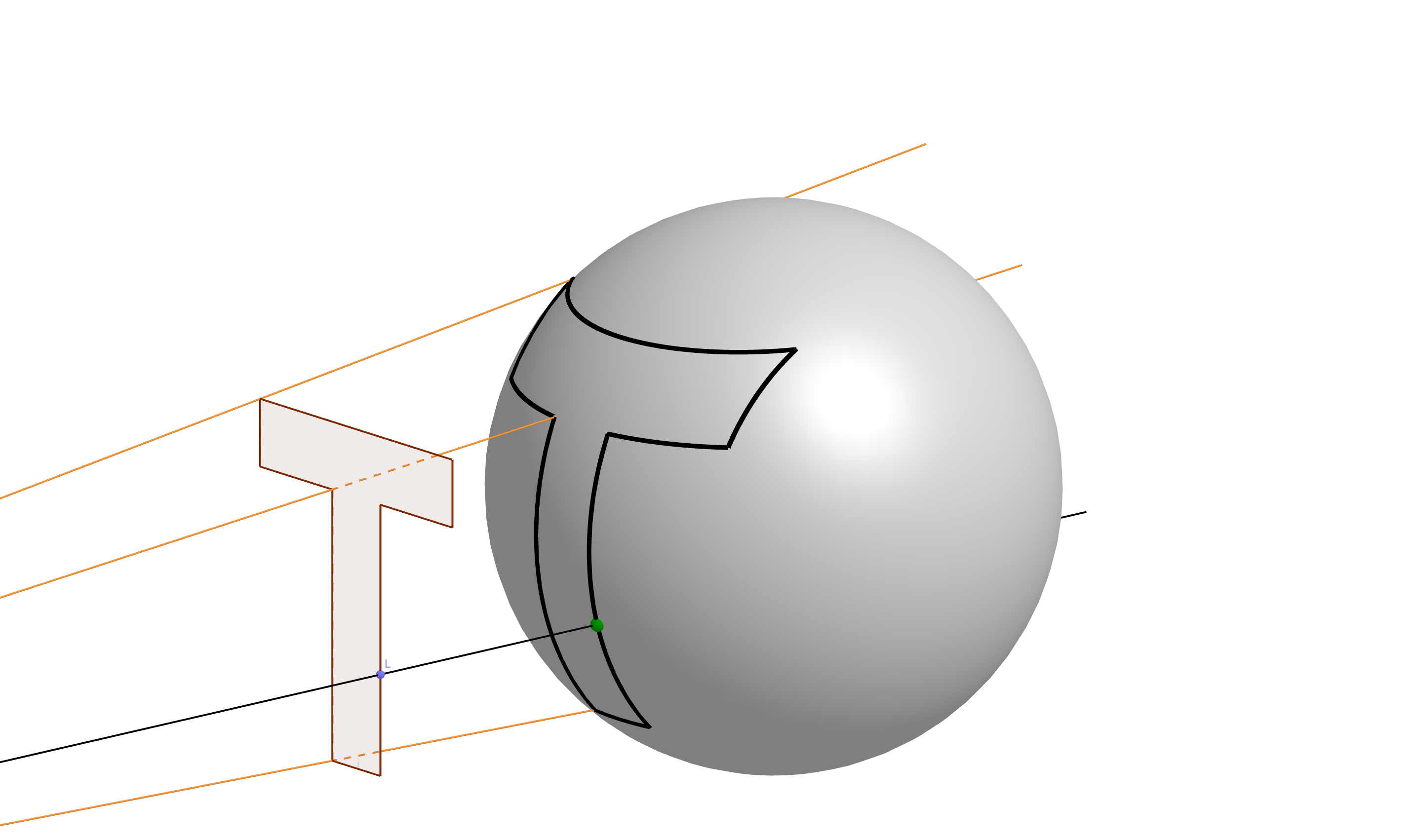}
		\caption{Projection of a letter onto the reachable space.}
		\label{fig:projection}
	\end{minipage}
	\vspace{-1.0em}
\end{figure}

The light painting task is set up as follows.
A piece of reflective tape is attached to the tip of the Furuta pendulum.
While the pendulum is moving, a long exposure photograph is taken.
The goal is to draw recognizable letters with the tip of the pendulum.
Since the reachable space of the Furuta pendulum covers a part of a sphere, as shown in~\cref{fig:reachable},
all letters first need to be projected onto the reachable space before drawing, as depicted in~\cref{fig:projection}.

Ideally, we would like to have a controller that receives a letter as input
and is able to trace its contour with the tip of the pendulum.
If the system was fully actuated, trajectory tracking would be straightforward.
However, due to underactuation, not every trajectory can be executed but only dynamically feasible ones.
Therefore, the crucial task is to find a trajectory which most closely follows the shape of the desired letter.
As a proxy for this task, we discretize the letter into a sequence of waypoints and subsequently search for a trajectory that passes through these waypoints.

\begin{figure}[t]
	\centering
	\includegraphics[width=0.98\linewidth]{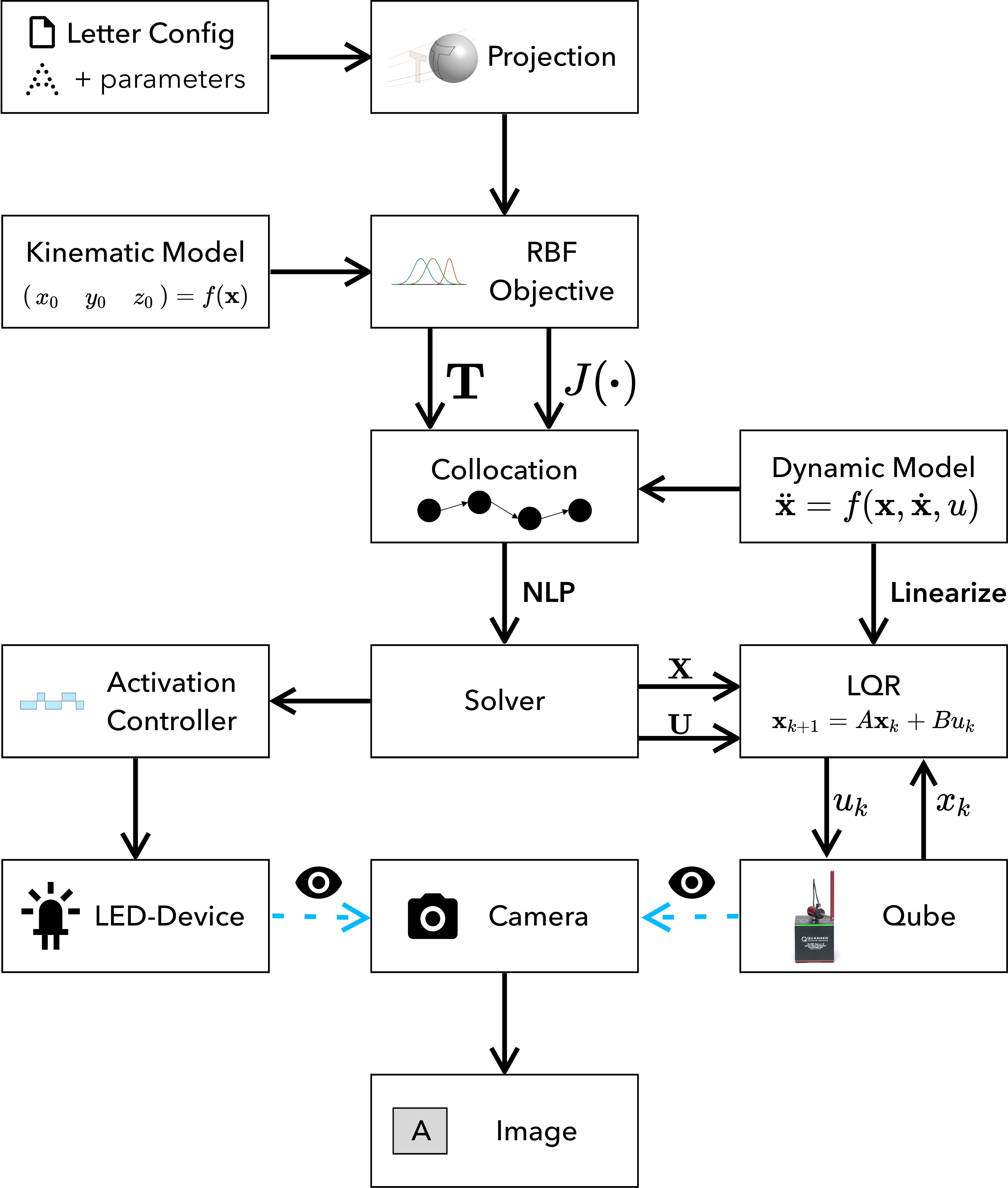}
	\caption{Full pipeline from planning to tracking for drawing letters
		via light painting photography using Furuta pendulum.}
	\label{fig:pipeline}
	\vspace{-1.0em}
\end{figure}

The diagram in~\cref{fig:pipeline} shows the full pipeline of our approach to underactuated light painting.
On a high level, it can be split into three parts, from top to bottom:
\emph{waypoint generation} (first row),
\emph{trajectory optimiztion} (\mbox{rows 2--3}),
and \emph{execution} (bottom 3 rows).
Waypoint generation comprises letter discretization and projection discussed above.
Trajectory optimization takes the generated waypoints as input
and finds a sequence of control commands that drives the system through these waypoints
using the knowledge of the system kinematics and dynamics.
Finally, an LQR feedback controller is added for tracking of the optimized trajectory at the execution stage.
Additionally, a synchronized set of LEDs is activated
when the pendulum passes through the trajectory segments belonging to the letter to increase illumination.

\section{Trajectory Optimization}
\label{sec:planning}
Given a set of waypoints obtained via letter discretization and subsequent projection onto the reachable space,
we aim to devise an objective function that will yield a trajectory passing through the waypoints upon optimization.
To this end, we first describe the trajectory optimization method which we employ in~\cref{subsec:dircol}.
After that, we present the main idea of our approach of introducing `attention' into the optimization objective
and explain it on the task of reaching a single desired waypoint in~\cref{subsec:single_point}.
Finally, in~\cref{subsec:multiple_points}, we demonstrate how the idea of introducing `attention' can be extended to multiple waypoints and how to enforce a desired ordering among them.

\subsection{Direct Collocation}
\label{subsec:dircol}
Trajectory optimization is concerned with finding a feasible trajectory that minimizes a given objective function.
Numerical optimization methods such as multiple shooting and direct collocation work by transforming a continuous-time
optimal control problem into a big \ac{nlp}~\cite{betts1998survey}.
Methods differ in how exactly the discretization is done and what variables are treated as optimization variables.
We use direct collocation with cubic splines~\cite{hargraves1987direct},
widely spread in robotics~\cite{tedrake2009underactuated},
and implement our optimization problem in CasADi~\cite{Andersson2018}.

Direct collocation treats both states $\mathbf{x}_t$ and control commands $\mathbf{u}_t$ as optimization variables,
\begin{equation*}
\mathbf{X} =
	\begin{bmatrix}
		\mathbf{x}_0 & \ldots & \mathbf{x}_{N}
	\end{bmatrix},
	\qquad
\mathbf{U} =
	\begin{bmatrix}
		\mathbf{u}_0 & \ldots & \mathbf{u}_{N-1}
	\end{bmatrix},
\end{equation*}
whereas the system dynamics are imposed as constraints.
The objective function typically has the form of a sum over the time steps
\begin{equation}
\label{eqn:alpha_hard}
	J(\mathbf{X}, \mathbf{U}) = \underbrace{\sum_{t=0}^{N} \alpha_t d(\hat{\mathbf{x}}, \mathbf{x}_t)}_{J_{\alpha}(\mathbf{X})} + 
	\underbrace{\beta \sum_{t=0}^{N-1} \mathbf{u}_t^2}_{J_{\beta}(\mathbf{U})}
\end{equation}
where $d(\hat{\mathbf{x}}, \mathbf{x}_t)$ is a distance-based metric that encodes the state-dependent part of the running cost.
Weights $\alpha_t \in [0, 1]$ determine the importance of each time step and are usually set to $\alpha_t = 1$.
Parameter $\beta$ is chosen such that the cost of the squared control commands is orders of magnitude smaller than the other cost terms.
Moreover, we introduce $\hat{\mathbf{x}}$ as a parameter, which will later play the role of a waypoint.

Our key idea is to parameterize the weights $\alpha_t$ in a specific way
that draws the `attention' of the optimizer to the important moments in time when the waypoints need to be reached.
Crucially, which moments exactly are important is determined by the optimizer itself.
In the following, we detail how this is done, first on a single-waypoint example and then on the full sequential problem.
\begin{figure}[t]
	\centering
	\includegraphics[width=1.0\linewidth]{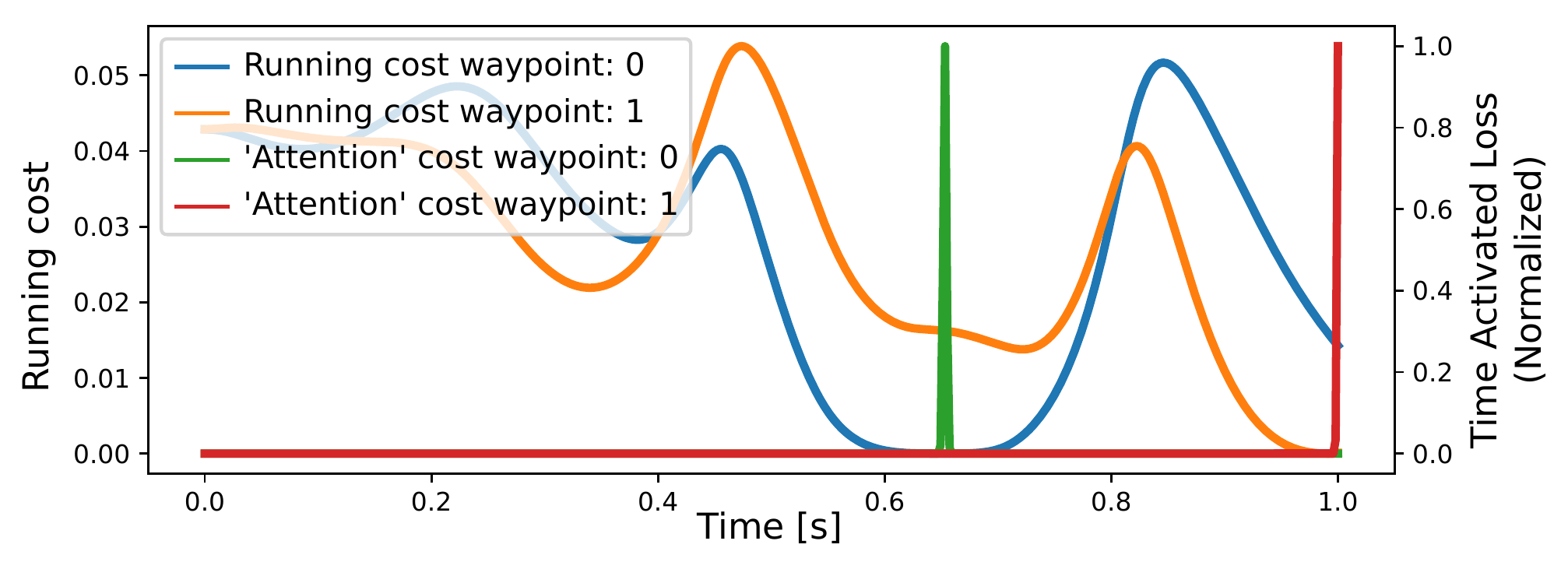}
	\caption{Effect of `attention' on the loss function.
		Components of the loss function are drawn over time for a task with two waypoints.
		Note that the loss is minimal and close to zero when `attention' is one.}
	\label{fig:two_via_dyn_activation}
	\vspace{-1.0em}
\end{figure}

\subsection{Attention Mechanism for Reaching a Single Waypoint}
\label{subsec:single_point}
If there is only one point $\hat{\mathbf{x}}$ that needs to be reached,
the coefficients $\alpha_t$ in~\cref{eqn:alpha_hard} can be set as
\begin{equation}
\label{eqn:alpha_single_point}
\alpha_t = \begin{cases}
1, & t = N, \\
0, & \text{otherwise},
\end{cases}
\end{equation}
which puts all the weight on the last time step  and yields a trajectory that ends up at the target state.
At first sight, one could imagine solving a set of such one-waypoint problems
and then chaining the solutions together to obtain a complete trajectory.
However, this approach will not work, because it does not account for the fact that the final state of one segment becomes the initial condition for the subsequent one.
Since the dynamics are nonlinear and the system is underactuated,
the optimizer may decide to e.g. do an additional swing between going from one waypoint to another, despite the points being next to each other, just because the velocity with which the first waypoint was reached was not sufficiently high.
Moreover, switching controllers between segments is non-trivial and leads to jerky transitions.
Therefore, we aim for developing a method that allows to pass through multiple waypoints smoothly instead.

An approach based on~\cref{eqn:alpha_single_point},
where the activation time is trivially set to the last time step,
is hardly scalable to multiple waypoints, as the activation time for each point would have to be known in advance.
Setting the activation times for multiple waypoints by hand is prohibitive and in general leads to suboptimal solutions.
This can be attributed to the underactuated and oscillating nature of the Furuta pendulum, which makes it hard to anticipate how much swinging is needed to accumulate sufficient energy for reaching certain states.

For long exposure photography, it does not matter at what exact time the system passes through each waypoint.
This renders hard-coded activations such as in~\cref{eqn:alpha_single_point} unnecessary
and motivates a more flexible approach.
Namely, instead of pre-specifying the activations $\alpha_t$, we treat them as optimization variables.
More concretely, we parameterize the coefficients $\alpha_t$ by \acp{rbf} of the form
\begin{equation}
\label{eqn:alpha_dyn}
\alpha_t = \exp \left(-\frac{(\hat{t} - t)^2}{\sigma^2}\right)
\end{equation}
where $\hat{t}$ is the center of the \ac{rbf} and $\sigma$ is the bandwidth.
The center $\hat{t}$ determines the activation time and is introduced as a new optimization variable in the \ac{nlp}.
Thus, the optimizer is able to shift its `attention'
and can account for the time needed to accumulate sufficient energy to reach a desired state.
Inserting~\cref{eqn:alpha_dyn} into~\cref{eqn:alpha_hard},
we obtain the objective function that incorporates `attention' for a single waypoint.
\begin{equation}
\label{eqn:alpha_dyn_single_point}
	J_{\alpha}(\mathbf{X}, \hat{t}) = 
	\sum_{t=0}^{N} \exp \left(-\frac{(\hat{t} - t)^2}{\sigma}\right) d(\hat{\mathbf{x}}, \mathbf{x}_t).
\end{equation}
To exclude trivial solutions achieved by shifting the attention out of the scope of the finite trajectory,
$\hat{t}$ needs to be constrained to the interval $[0, N]$.

This formulation also allows one to minimize the time of arrival at the waypoint
by simply adding a punishment term $\gamma \hat{t}$ to the objective function in~\cref{eqn:alpha_dyn_single_point}
with some positive weight $\gamma$.
The main advantage of this approach is its independence on pre-specified activation times,
which also makes it scalable to multiple waypoints.

\subsection{Attention for Reaching Multiple Waypoints in Sequence}
\label{subsec:multiple_points}
Extending~\cref{eqn:alpha_dyn_single_point} with an activation time $\hat{t}_i$
for each waypoint $\hat{\mathbf{x}}_i$ and summing over the waypoints,
we obtain the objective function for multiple waypoints
\begin{equation}
\label{eqn:dyn_multi_via_loss}
J_{\alpha}(\mathbf{X}, \mathbf{T}) = 
	\sum_{i=0}^{M-1}\sum_{t=0}^{N} \exp \left( -\frac{(\hat{t}_i - t)^2}{\sigma} \right)
	d(\hat{\mathbf{x}}_i, \mathbf{x}_t)
\end{equation}
where $M$ is the number of waypoints and
$\mathbf{T}$ is the set of their associated activation times $\hat{t}_i$.

The order in which the waypoints are traversed matters:
if the waypoints are traversed in an arbitrary order, the drawn letters are hardly recognizable.
However, the ordering is not enforced by the objective function in~\cref{eqn:dyn_multi_via_loss}.
To impose order, we augment the optimization problem with constraints of the form
$\hat{t}_i \leq \hat{t}_{i+1},\, i =0, \ldots, M-2$.
Furthermore, it is beneficial to split up the set of the waypoints into segments.
All segments are then treated within one \ac{nlp},
but the ordering constraints are only enforced within each segment.
To further improve the smoothness of the trajectory,
we add a punishment term
\begin{equation}
\label{eqn:segment_duration}
J_\mu(\mathbf{T}) = \mu \sum_{j=0}^{S-1} \left(\hat{t}_{l_j} - \hat{t}_{f_j}\right)
\end{equation}
to the objective function that favors short segments $\hat{t}_{l_j} - \hat{t}_{f_j}$.
Here, $f_j$ and $l_j$ denote the first and last waypoints in segment $j$, respectively.
Each letter is split into $S$ segments and $\mu$ determines
the strength of the segment duration punishment.
The resulting objective function for multiple segments is given by
\begin{equation}
\label{eqn:objective}
J(\mathbf{X}, \mathbf{U}, \mathbf{T}) = 
	J_\alpha(\mathbf{X}, \mathbf{T}) + J_\beta(\mathbf{U}) + J_\mu(\mathbf{T}).
\end{equation}
The \ac{nlp} is then solved by minimizing the objective~\cref{eqn:objective}
subject to the collocation constraints on the system dynamics, path and boundary constraints,
and the proposed activations ordering constraints.

\cref{fig:two_via_dyn_activation} illustrates the effect of `attention' on the loss function.
The results were obtained by optimizing the objective given in~\cref{eqn:dyn_multi_via_loss}
with two waypoints.
The curves correspond to the individual terms $d(\hat{\mathbf{x}}_i, \mathbf{x}_t)$ and~$\alpha_t^i$
for each of the two waypoints $i=0$ and $i=1$.
Thus, the value of the full loss is given by the sum of the terms
$\alpha_t^i d(\hat{\mathbf{x}}_i, \mathbf{x}_t)$ over \mbox{$t$ and $i$}.
Notably, when `attention' rises to one, the corresponding distance-based loss goes to zero,
signalling that the waypoint is reached.
Summing the losses up without time activations would yield a high value for the total cost,
despite both waypoints being reached (indicated by the loss going to zero once for each waypoint).
Therefore, a formulation with a flat weighting $\alpha_t = 1$ for all time steps,
as it is used in most of the literature, would yield a high loss value despite the desired states being reached.
In contrast, the \ac{rbf}-based objective function in~\cref{eqn:dyn_multi_via_loss},
which only accumulates the distance-based losses close to the waypoints, results in a much lower loss value.

\section{Linear-Quadratic Optimal Tracking}
\label{sec:tracking}
\begin{figure}[t]
	\centering
	\includegraphics[width=\linewidth]{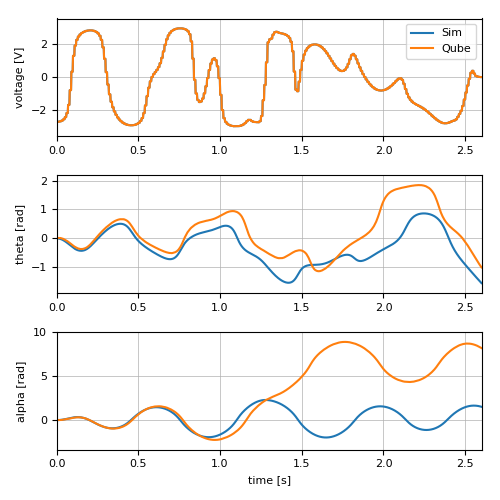}
	\caption{Open-loop control on the Quanser Qube:
		(top) motor input voltage,
		(middle) horizontal joint angle $\theta$,
		(bottom) pendulum joint angle $\alpha$,
		plotted over time.
		Trajectory in blue (Sim) was optimized in simulation to trace
		letter `S' as shown in~\cref{fig:sim_trajectories}.
		Trajectory in orange (Qube) was obtained on the real system,
		and it rather quickly diverges from simulation.
	}
	\label{fig:openloop}
	\vspace{-1.0em}
\end{figure}
\begin{figure}[t]
	\centering
	\includegraphics[width=\linewidth]{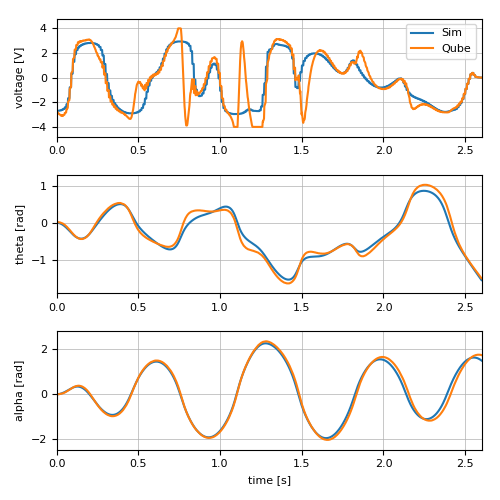}
	\caption{Closed-loop LQR tracking controller successfully tracks the same desired trajectory
		as in \cref{fig:openloop}, and it is able to correct the deviations from the planned trajectory
		despite underactuation.
		Undesirable overshoots in the input voltage, owing to exploits of the linearized system dynamics,
		are clipped to prevent excessively large control signals.
	}
	\label{fig:tracking}
	\vspace{-1.0em}
\end{figure}
Executing an open-loop sequence of control commands on the real system
results in a trajectory rather quickly diverging from the desired path
due to disturbances, modeling errors, and uncertainties in the initial conditions.
An example is shown in~\cref{fig:openloop}.
To prevent such divergence and to keep the system on the desired trajectory,
we employ an LQR tracking controller described in the following.

The first step is to linearize the system dynamics along a desired trajectory.
If $\dot{\mathbf{x}} = \mathbf{f}(\mathbf{x}, \mathbf{u})$,
then the linearization around a given point $(\mathbf{x}^d, \mathbf{u}^d)$ can be written as
\begin{equation} \label{eq:taylor}
\dot{\mathbf{x}} \approx \dot{\mathbf{x}}^d + \frac{\partial \mathbf{f}(\mathbf{x}^d, \mathbf{u}^d)}{\partial \mathbf{x}} (\mathbf{x} - \mathbf{x}^d) + \frac{\partial \mathbf{f}(\mathbf{x}^d, \mathbf{u}^d)}{\partial \mathbf{u}} (\mathbf{u} - \mathbf{u}^d).
\end{equation}
Performing such linearization at every time step,
we can obtain a linearization around the desired trajectory.
It is convenient to introduce auxiliary variables
representing the deviations from the desired trajectory
\begin{equation}
\label{eq:coordinate}
\tilde{\mathbf{x}}_t = \mathbf{x}_t - \mathbf{x}_t^d, \qquad
\tilde{\mathbf{u}}_t = \mathbf{u}_t - \mathbf{u}_t^d.
\end{equation}
Inserting~\cref{eq:coordinate} into~\cref{eq:taylor}, we obtain
\begin{equation} \label{eq:lin_cont}
\dot{\tilde{\mathbf{x}}}_t = \frac{\partial \mathbf{f} (\mathbf{x}_t^d, \mathbf{u}_t^d)}{\partial \mathbf{x}} \tilde{\mathbf{x}}_t + \frac{\partial \mathbf{f} (\mathbf{x}_t^d, \mathbf{u}_t^d)}{\partial \mathbf{u}} \tilde{\mathbf{u}}_t.
\end{equation}
Discretizing the continuous-time linear dynamical system given in~\cref{eq:lin_cont} using the Euler integration scheme
\begin{equation} \label{eq:discrete_dynamics}
\tilde{\mathbf{x}}_{t+1} = \tilde{\mathbf{x}}_t + \Delta t \dot{\tilde{\mathbf{x}}}_t,
\end{equation}
we arrive at the discrete-time time-varying dynamics
\begin{align} \label{dls}
\tilde{\mathbf{x}}_{t+1}
&= \left(
		\mathbf{I} + \Delta t \frac{\partial \mathbf{f} (\mathbf{x}_t^d, \mathbf{u}_t^d)}{\partial \mathbf{x}}
	\right)
	\tilde{\mathbf{x}}_t
+ \Delta t \frac{\partial \mathbf{f} (\mathbf{x}_t^d, \mathbf{u}_t^d)}{\partial \mathbf{u}}
	\tilde{\mathbf{u}}_t \nonumber \\
&= \mathbf{A}_t \tilde{\mathbf{x}}_t + \mathbf{B}_t \tilde{\mathbf{u}}_t.
\end{align}
These dynamics provide the basis for designing a time-varying tracking feedback controller.

Given the linearized model along the trajectory in~\cref{dls},
we can formulate the trajectory stabilization problem as the minimization of the cost
\begin{equation}
	\label{eq:lqr_cost}
J = \sum_{t=0}^{N-1} \left(
		\tilde{\mathbf{x}}_t^T \mathbf{Q} \tilde{\mathbf{x}}_t + \tilde{\mathbf{u}}_t^T \mathbf{R} \tilde{\mathbf{u}}_t
	\right).
\end{equation}
The system is quadratically penalized for being away from the desired trajectory
using weighting matrices $\mathbf{Q}$ and $\mathbf{R}$.
The optimal feedback controller that minimizes the cost given in~\cref{eq:lqr_cost}
subject to the dynamics provided in~\cref{dls} is an affine control law of the form
\begin{equation}
	\label{eq:lqr_ctrl}
\mathbf{u}_t = \mathbf{u}_t^d - \mathbf{K}_t \tilde{\mathbf{x}}_t
\end{equation}
where the feedback gain matrix $\mathbf{K}_t$ is found by solving the discrete-time Riccati equation
backwards in time~\cite{lewis2012optimal}.

The result of applying the stabilizing LQR controller derived in~\cref{eq:lqr_ctrl}
to the same trajectory on which the open-loop execution failed is shown in~\cref{fig:tracking}.
As it can be seen from the plots, the system is able to follow the desired trajectory, canceling all disturbances and deviations, in spite of being underactuated.

Notwithstanding its impressive performance,
the LQR as a tracking controller for the Furuta pendulum has some limitations.
First, the controller can only stabilize the system when it is sufficiently close to the desired trajectory.
Due to underactuation, the envelope of correctable deviations is quite small.
Second, due to high nonlinearity of the dynamics, linearizations can be rather bad in some states,
leading to overshooting and instability.
As the LQR has no natural way of incorporating control constraints, the applied control voltages were clipped.

Another general problem of the LQR is the choice of the weighting matrices $\mathbf{Q}$ and $\mathbf{R}$,
which are typically found using prior knowledge or trial-and-error.
We were able to find good parameters for the presented examples,
but as generated trajectories for different letters show significant variability,
a tailored set of parameters is required for each letter.
A similar problem is stated in~\cite{divelbiss1997trajectory}.
Finding a good set of parameters without many trials is still an open research area
and could be the subject for future work,
potentially solved by learning or optimization algorithms such as~\cite{marco2016automatic}.

\section{Results}
\label{sec:results}
\begin{figure}[t]
	\centering
	\begin{subfigure}{0.41\columnwidth}
		\centering
		\includegraphics[width=\linewidth]{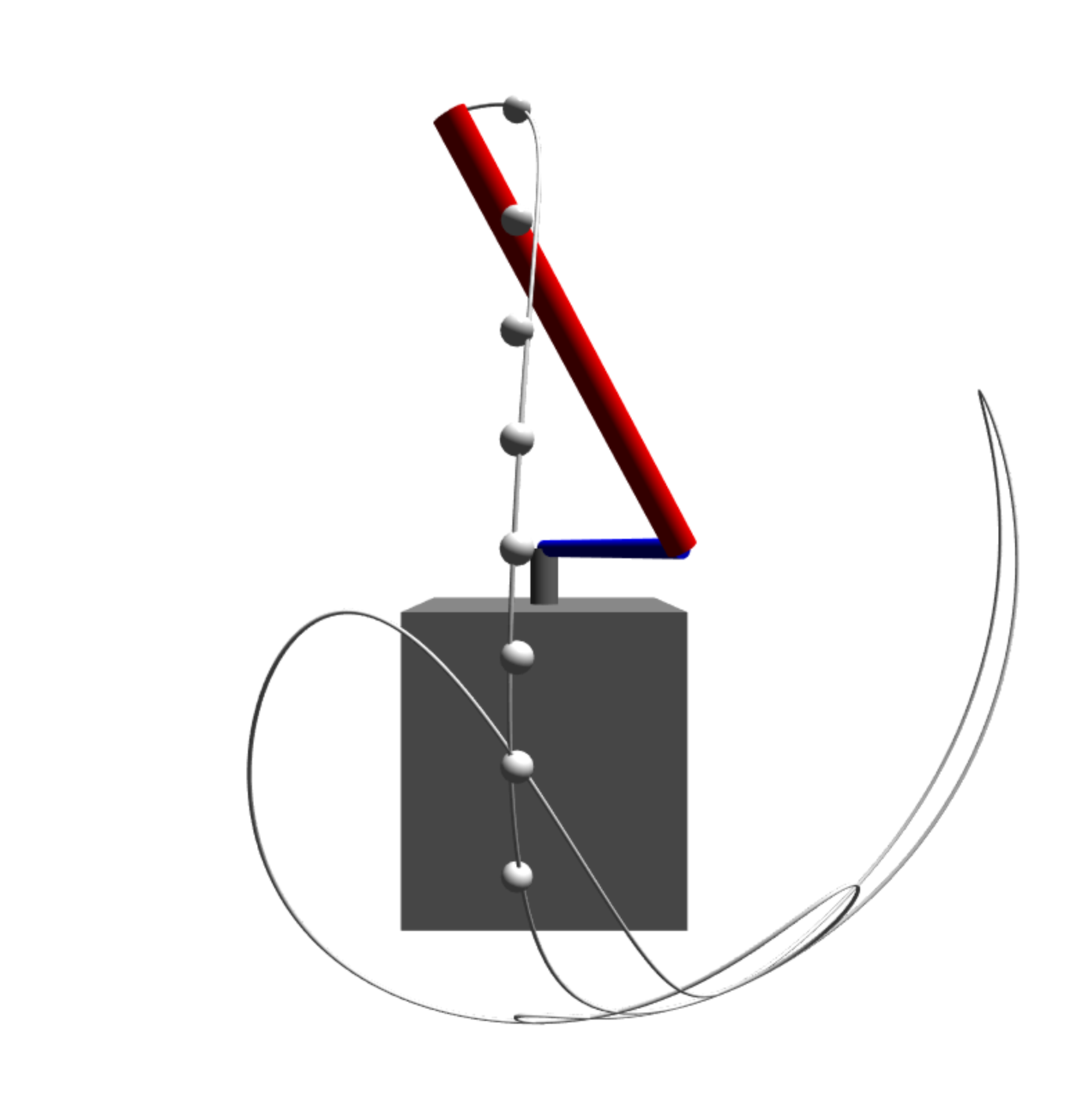}
	\end{subfigure}\quad
	\begin{subfigure}{0.41\columnwidth}
		\centering
		\includegraphics[width=\linewidth]{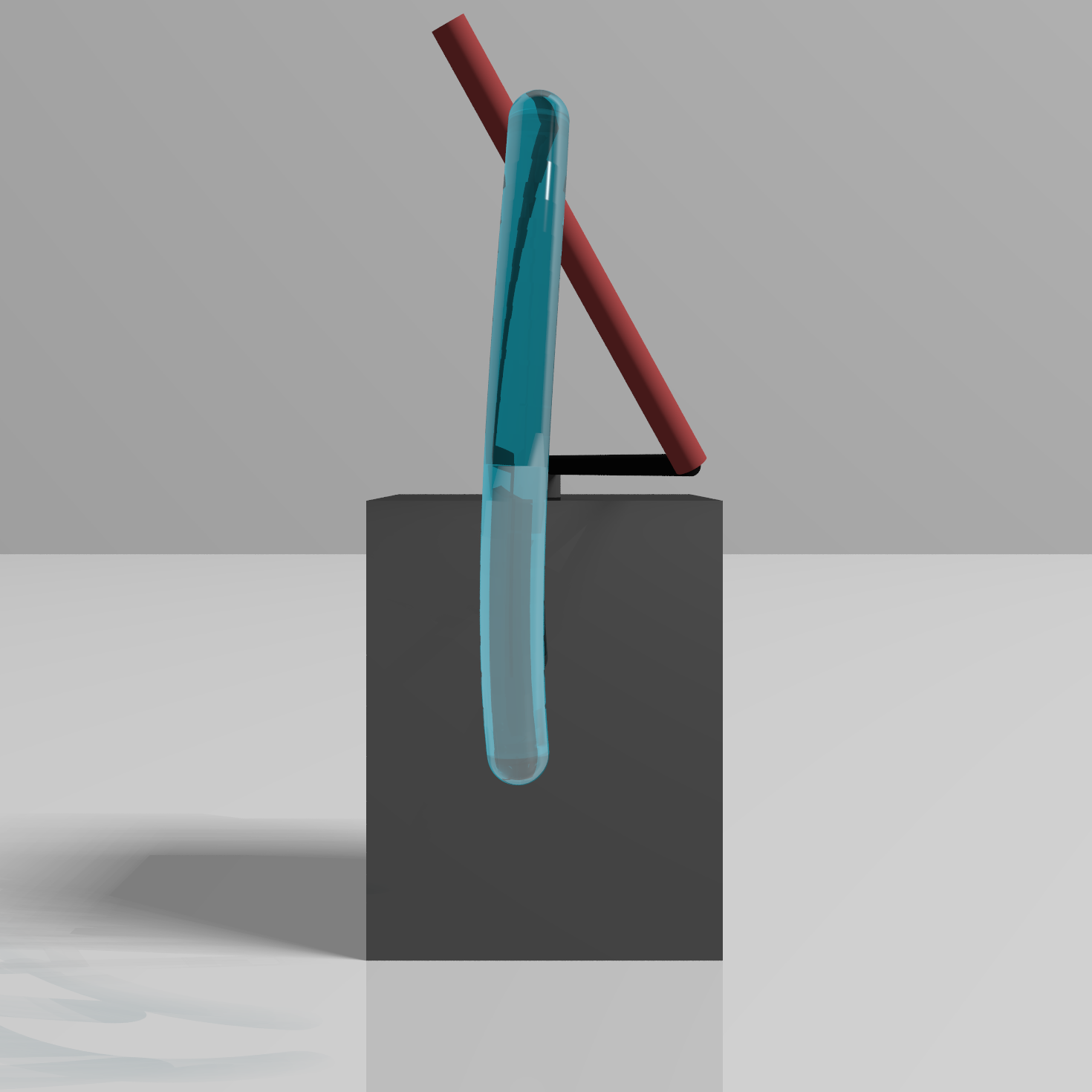}
	\end{subfigure}
	\begin{subfigure}{0.41\columnwidth}
		\centering
		\includegraphics[width=\linewidth]{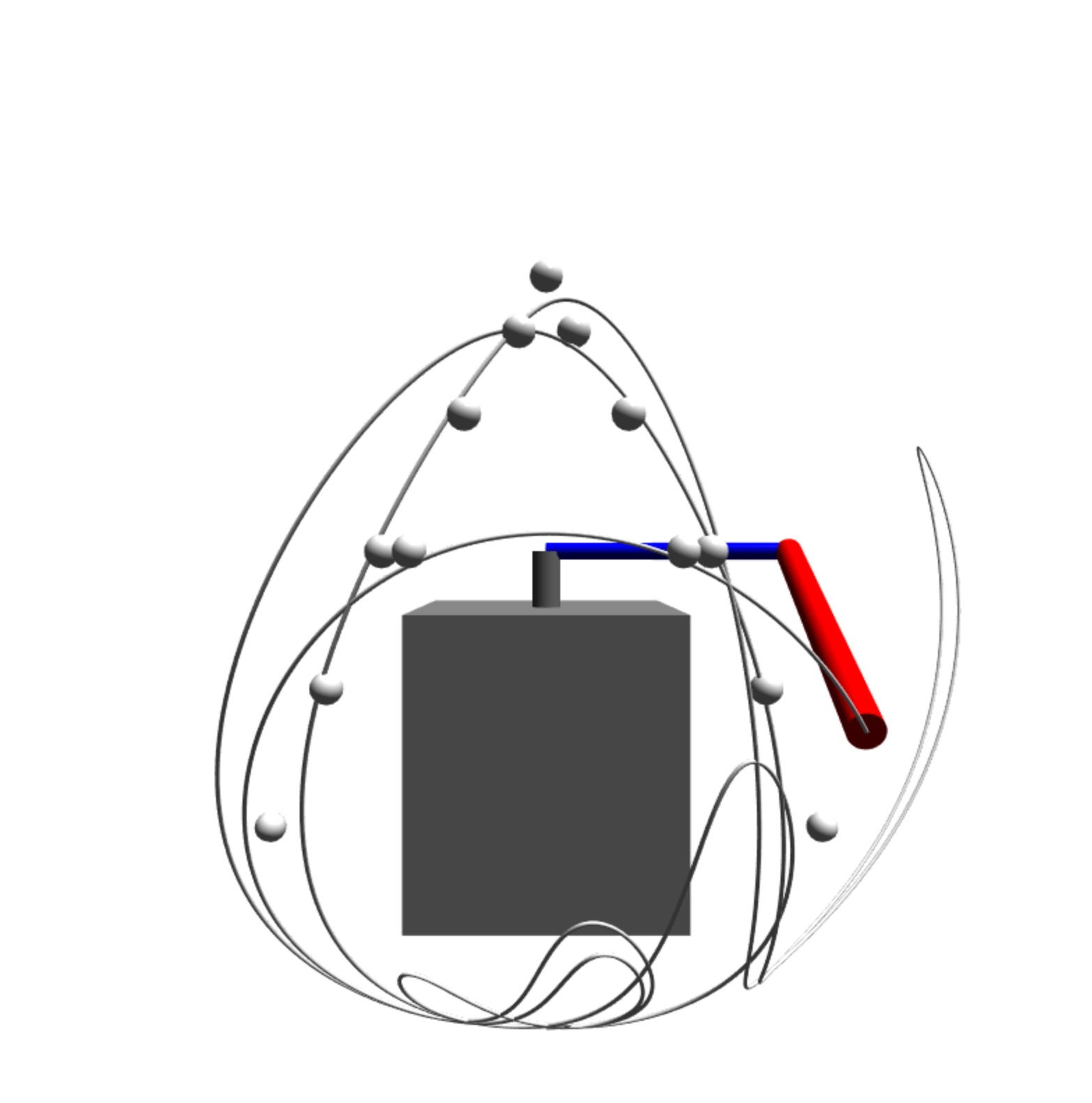}
	\end{subfigure}\quad
	\begin{subfigure}{0.41\columnwidth}
		\centering
		\includegraphics[width=\linewidth]{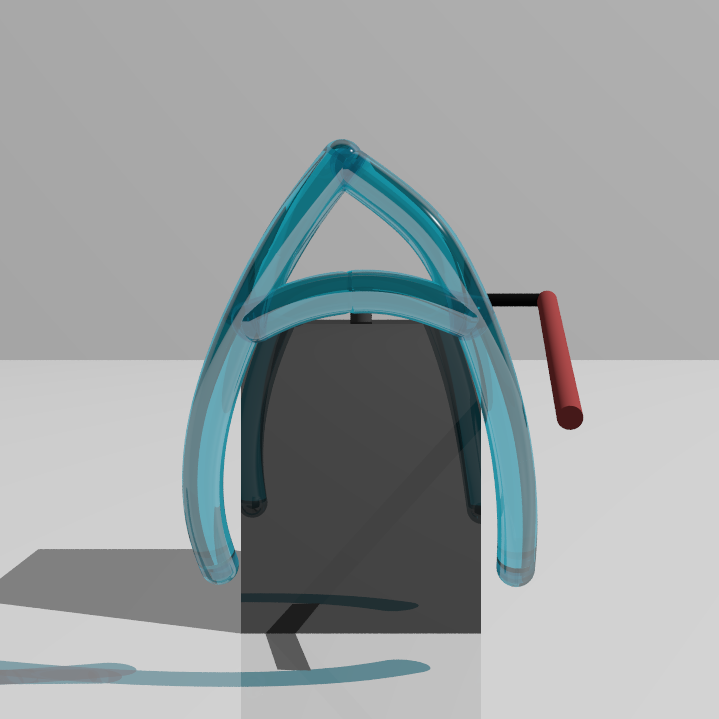}
	\end{subfigure}
	\begin{subfigure}{0.41\columnwidth}
		\centering
		\includegraphics[width=\linewidth]{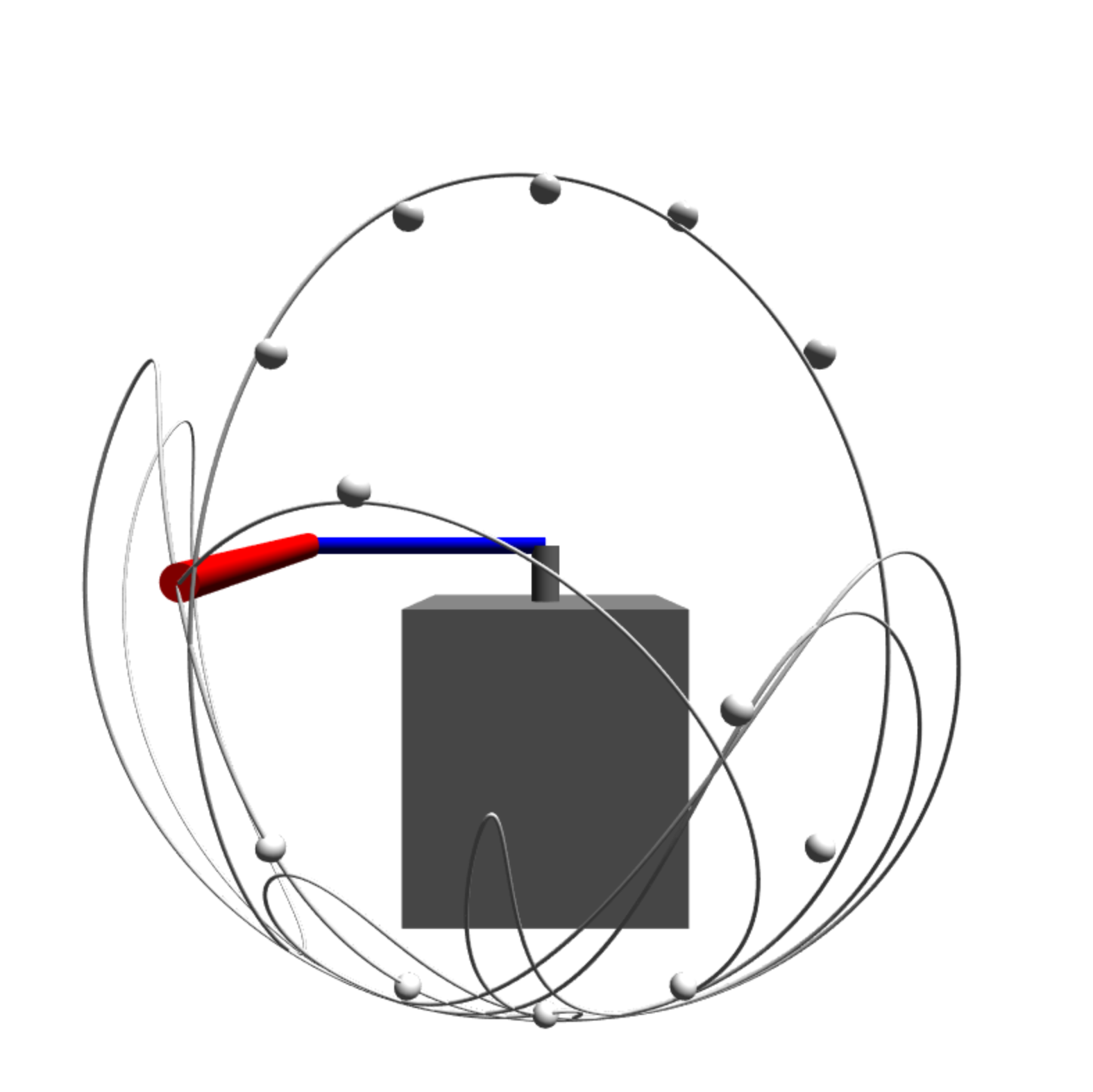}
		\caption{Trajectory trace.}
	\end{subfigure}\quad
	\begin{subfigure}{0.41\columnwidth}
		\centering
		\includegraphics[width=\linewidth]{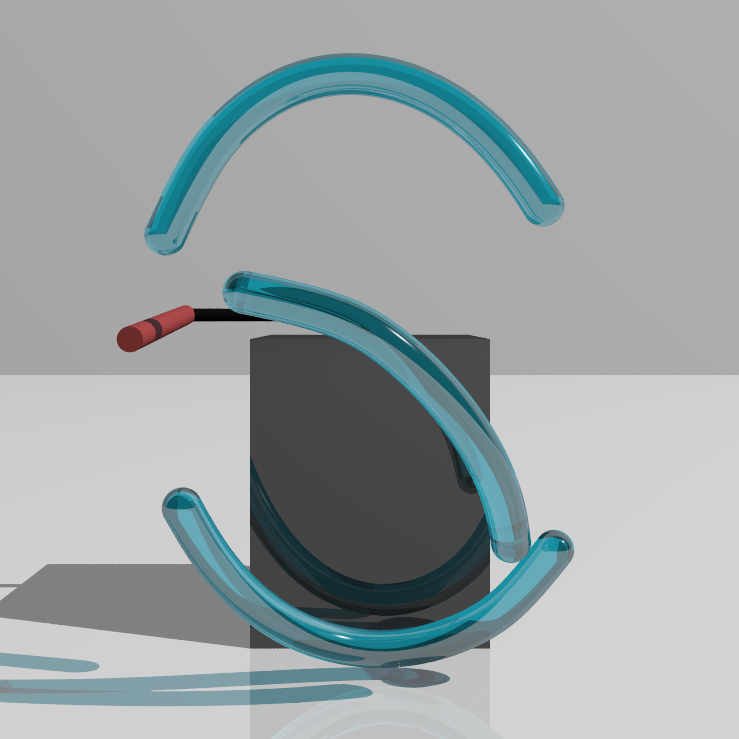}
		\caption{Light painted shape.}
	\end{subfigure}
	\caption{Pendulum trajectories and corresponding light painted letter shapes.
		Each highlighted segment on the right consists of a set of waypoints traversed in quick succession.
		Note that the complete trajectories may be quite long, as seen in the traces on the left,
		and it is virtually impossible to design such trajectories by hand or using kinematic path planning.
	}
	\label{fig:sim_trajectories}
	\vspace{-1.0em}
\end{figure}
In the previous sections, individual blocks from the pipeline in~\cref{fig:pipeline} have been introduced.
In this section, the complete approach is evaluated and the resulting light painted trajectories are presented.

The Quanser Qube implementation of the Furuta pendulum imposes a hard limit on the range of values that the horizontal rotary angle $\theta$ can take, reflected in the reachable space shown in~\cref{fig:reachable}.
In addition, a software limit is imposed on the input voltage signal $u$ to avoid damaging the motor.
To account for the joint and control limits, the following inequality constraints
\begin{equation*}
- u_{\text{max}} \leq u_t \leq u_{\text{max}}, \qquad - \theta_{\text{max}} \leq \theta_{t} \leq \theta_{\text{max}}
\end{equation*}
are added to our direct collocation formulation of the trajectory optimization problem described in~\cref{sec:planning}.

For all of our experiments, the initial state $\mathbf{x}_0$ is assumed to be zero, which corresponds to the system being still, with the pole centered in the front and hanging down.

Following the pipeline from~\cref{fig:pipeline}, the
trajectories shown in~\cref{fig:sim_trajectories} were obtained in simulation.
The traces on the left show that a significant amount of time is spent in preparation of each maneuver,
while the pendulum is accumulating the required energy and momentum to pass through the waypoints
in the specified order and in quick succession.
The visualizations on the right show the expected results from the light painting photography,
where the letter segments are highlighted based on the activation times $\hat{t}_i$
obtained through the `attention'-augmented trajectory optimization described \mbox{in~\cref{subsec:multiple_points}}.
Note that while letter `I' consists of a single segment, letters `A' and `S' are comprised of three segments each. The letter `S' is specially challenging because of the kinodynamic structure of the Furuta pendulum.

\begin{figure}[t]
	\centering
	\begin{subfigure}{0.32\columnwidth}
		\centering
		\includegraphics[width=\linewidth]{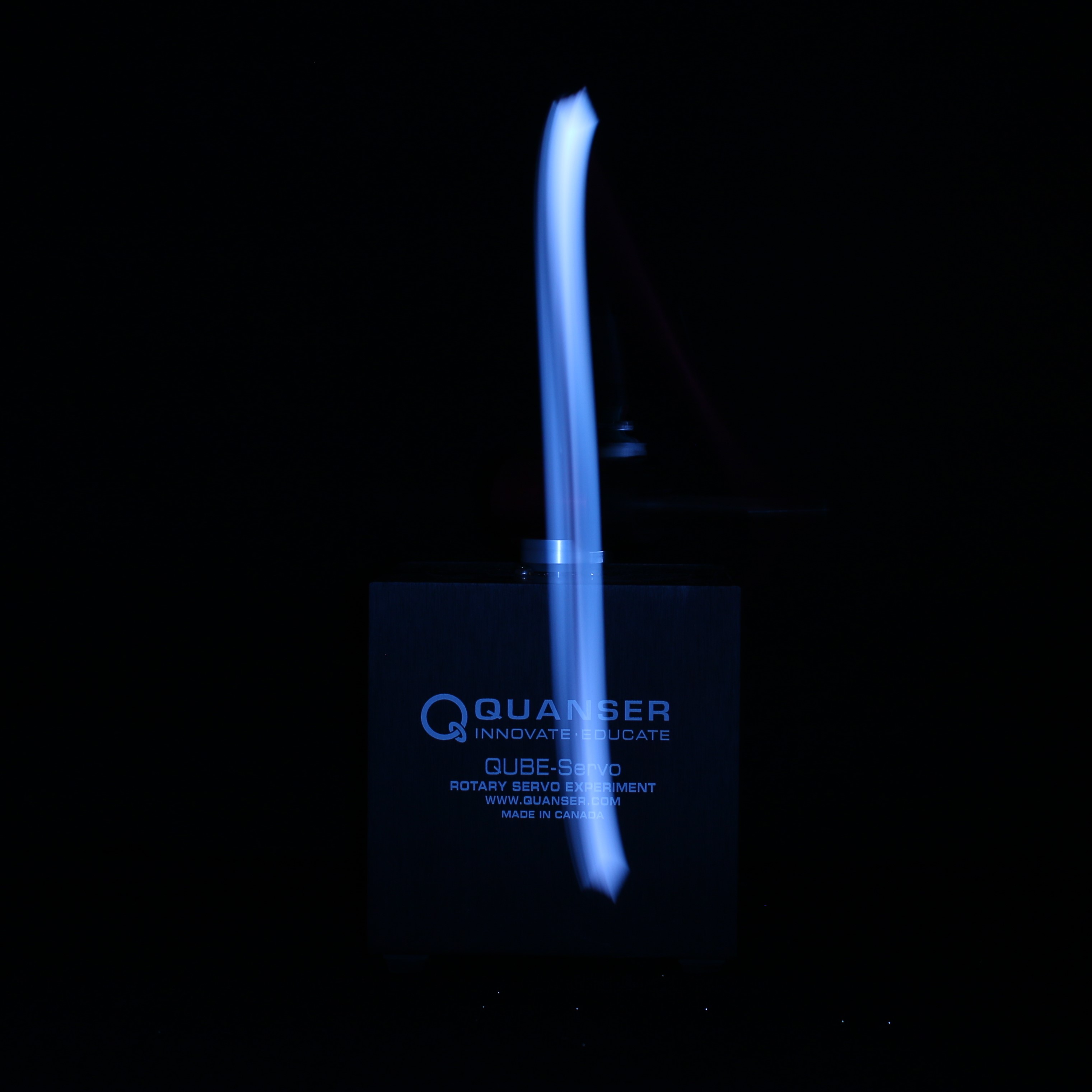}
	\end{subfigure}
	\begin{subfigure}{0.32\columnwidth}
		\centering
		\includegraphics[width=\linewidth]{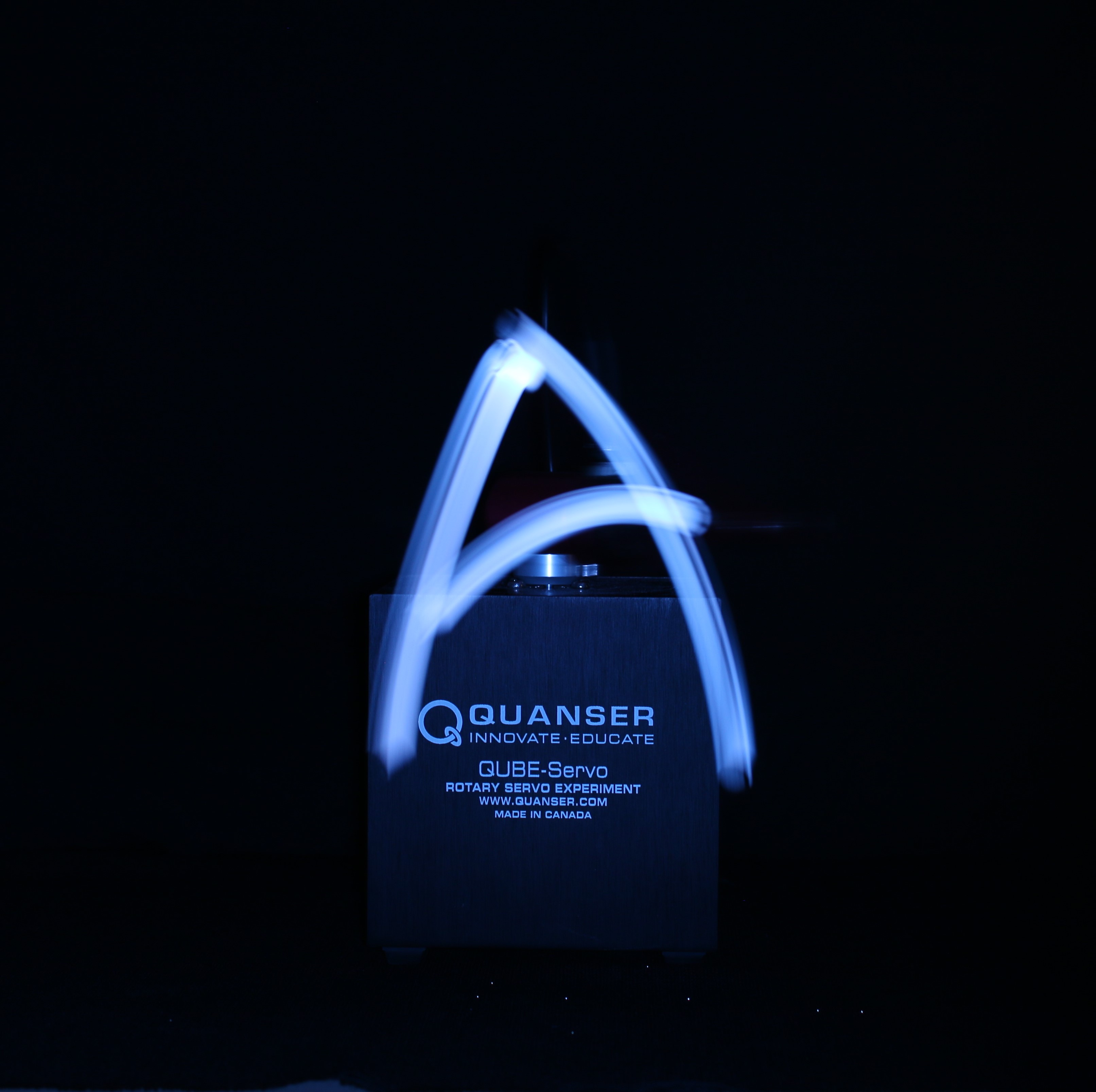}
	\end{subfigure}
	\begin{subfigure}{0.32\columnwidth}
		\centering
		\includegraphics[width=\linewidth]{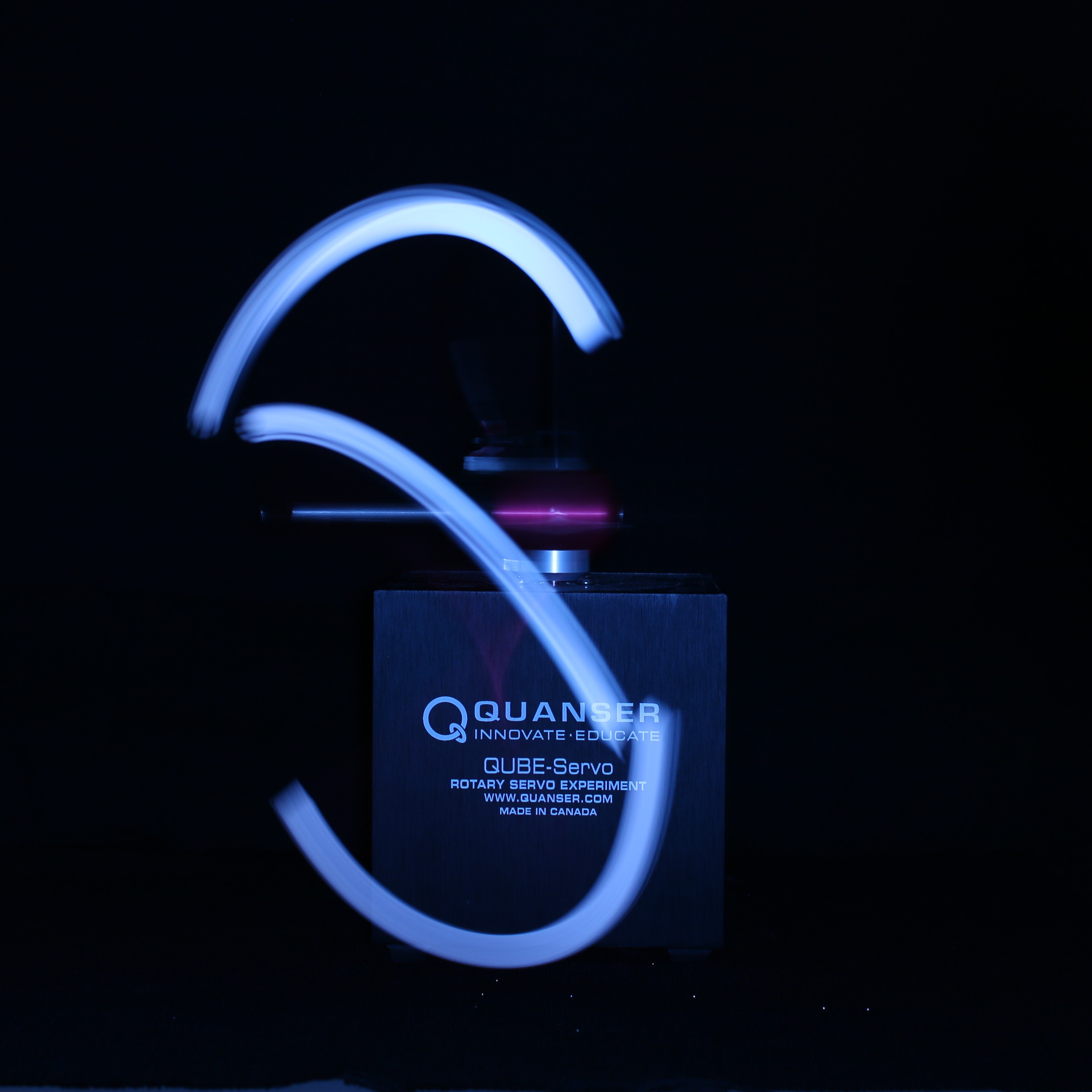}
	\end{subfigure}
	\caption{Images of letters `I', `A', and `S' created by light painting photography
		following the pipeline from~\cref{fig:pipeline}.}
	\label{fig:lep}
	\vspace{-1.0em}
\end{figure}

Long exposure photographs of the light painted letters are presented in~\cref{fig:lep}.
The pictures have been taken in a dark room with an LED device synchronized with the trajectory execution and activated based on the optimized segment beginning/end times $\hat{t}_i$
described \mbox{in~\cref{subsec:multiple_points}}.
Comparing the real images in~\cref{fig:lep} with the simulated renderings in~\cref{fig:sim_trajectories},
we observe a sufficiently good match allowing the letters to be well recognizable.
However, the trajectories slightly deviate towards the end,
as it can be seen on the middle strokes in the letters `A' and `S' that are drawn last. These segments are slightly tilted compared to their desired location.
For a better view, see the accompanying video, where real and simulated trajectories are drawn side by side.

\section{Discussion and Conclusion}
\label{sec:conclusion}
A method for objective function design
in the context of trajectory optimization with waypoints has been presented (see~\cref{sec:planning}).
The proposed objective function (see~\cref{eqn:dyn_multi_via_loss}) features an RBF-smoothed `attention' over time that activates the distance-based loss when the corresponding waypoint is near.
Crucially, the RBF-activations are not hand-designed but jointly optimized together with the states and control commands.
For the tasks in which the order of the waypoints matters, the objective function has been extended to enforce the desired ordering (see~\cref{eqn:objective}).

The proposed method has been evaluated on a task of drawing letters with the Furuta pendulum, a highly dynamic underactuated system (see~\cref{sec:results}).
The letters were discretized into a set of waypoints, and a trajectory passing through them was optimized using the proposed objective function.
This procedure yielded activation times at which the waypoints were reached as a byproduct (see~\cref{fig:two_via_dyn_activation}).
An LQR-based tracking controller has been applied to execute the planned trajectories (see~\cref{sec:tracking}).
To visualize the trajectory traces, long exposure photography has been employed, with an LED ring illuminating the scene at the activation times obtained through optimization (see~\cref{fig:lep}).

Although the desired performance has been achieved, several improvements are possible.
First, the letter segmentation and discretization process should be automated.
Second, the complexity of waypoint optimization needs to be evaluated in more depth;
we used between \mbox{$5$ and $15$} waypoints, but larger numbers may be required in other tasks.
Finally, parameters such as time horizon, waypoints order, segment duration penalty, as well as the tracking LQR cost matrices are currently set by hand for each letter.
Automating this procedure would be of great practical interest even beyond the light painting task.

\section*{Appendix}
\label{appendix:eom}
Equations of motion of the Quanser Qube are given by
\begin{align*}
& \left(m_p L_r^2 + \frac{1}{4} m_p L_p^2 - \frac{1}{4} m_p L_p^2 \cos^2\alpha + J_r\right) \ddot{\theta} \\
& + \left(\frac{1}{2} m_p L_p L_r \cos\alpha\right) \ddot{\alpha} +
    \left(\frac{1}{2} m_p L_p^2 \sin\alpha \cos\alpha\right) \dot{\theta} \dot{\alpha} \\
& - \left(\frac{1}{2} m_p L_p L_r \sin\alpha\right) \dot{\alpha}^2 + D_r \dot{\theta}
	= \frac{k_m (u - k_m \dot{\theta})}{R_m}, \\
& \left(\frac{1}{2} m_p L_p L_r \cos\alpha \right) \ddot{\theta} +
    \left(J_p + \frac{1}{4} m_p L_p^2 \right) \ddot{\alpha} + D_p \dot{\alpha} \\
& - \left( \frac{1}{4} m_p L_p^2 \cos\alpha \sin\alpha \right) \dot{\theta}^2
	+ \frac{1}{2} m_p L_p g \sin\alpha = 0.
\end{align*}
The control command $u$ (see upper Eq.) is the motor voltage.
The dynamics parameters can be found in~\cite{qube-servo}.



\bibliographystyle{IEEEtran}
\bibliography{IEEEabrv,literature}

\end{document}